\pdfoutput=1

\documentclass[11pt]{article}

\usepackage[]{emnlp2021}

\usepackage{times}
\usepackage{latexsym}

\usepackage[T1]{fontenc}

\usepackage[utf8]{inputenc}

\usepackage{microtype}

\usepackage{amsmath}
\usepackage{algpseudocode}

\usepackage{graphicx}
\graphicspath{ {figures/} }
\usepackage{multirow}
\usepackage{tabularx}
\usepackage{hhline}
\usepackage{arydshln}
\usepackage{url}
\usepackage{enumitem}
\usepackage{booktabs}
\usepackage{xcolor}
\usepackage{setspace}
\usepackage{hyperref}
\usepackage{titlesec}
\usepackage{capt-of}%
\usepackage{varwidth}
\usepackage{wrapfig}
\usepackage{ctable}
\usepackage{adjustbox}
\usepackage{float}
\usepackage{subfiles}
\usepackage{amssymb}
\usepackage{enumitem}
\usepackage{colortbl}
\usepackage{array}
\usepackage{cleveref}
\usepackage{titlesec}
\usepackage{subfig}
\usepackage{lipsum}

\patchcmd{\footnotemark}{\stepcounter{footnote}}{\refstepcounter{footnote}}{}{}

\crefformat{section}{\S#2#1#3}
\crefformat{subsection}{\S#2#1#3}
\crefformat{subsubsection}{\S#2#1#3}
\crefrangeformat{section}{\S\S#3#1#4 to~#5#2#6}
\crefmultiformat{section}{\S\S#2#1#3}{ and~#2#1#3}{, #2#1#3}{ and~#2#1#3}
\crefformat{footnote}{#2\footnotemark[#1]#3}

\definecolor{darkpastelgreen}{rgb}{0.01, 0.75, 0.24}
\definecolor{darkterracotta}{rgb}{0.8, 0.31, 0.36}

\usepackage{microtype}

\usepackage{subfiles} 

\newcommand{\CDLM}{\textsc{CDLM}}

\definecolor{cadmiumorange}{rgb}{0.93, 0.53, 0.18}
\definecolor{darkspringgreen}{rgb}{0.09, 0.45, 0.27}
\definecolor{chamoisee}{rgb}{0.63, 0.47, 0.35}
\title{CDLM: Cross-Document Language Modeling}

\author{Avi Caciularu$^{1}\thanks{\;\; Work partly done as an intern at AI2.}$\hspace{1em} Arman Cohan$^{2,3}$\hspace{1em} Iz Beltagy$^{2}$\\
\textbf{Matthew E. Peters$^{2}$\hspace{1em} Arie Cattan$^{1}$\hspace{1em} Ido Dagan$^1$} \vspace{6pt}\\  
    $^1$Computer Science Department, Bar-Ilan University, Ramat-Gan, Israel\\
    $^2$Allen Institute for Artificial Intelligence, Seattle, WA\\
    $^3$Paul G. Allen School of Computer Science \& Engineering, University of Washington\\
    {\small\tt avi.c33@gmail.com, \{armanc,beltagy,matthewp\}@allenai.org} \\
    {\small\tt arie.cattan@gmail.com, dagan@cs.biu.ac.il }
}

\date{}
\begin{document}
\maketitle

\begin{abstract}
We introduce a new pretraining approach geared for multi-document language modeling, incorporating two key ideas into the masked language modeling self-supervised objective. First, instead of considering documents in isolation, we pretrain over sets of multiple related documents, encouraging the model to learn cross-document relationships. Second, we improve over recent long-range transformers by introducing dynamic global attention that has access to the entire input to predict masked tokens. We release CDLM (Cross-Document Language Model), a new general language model for multi-document setting that can be easily applied to downstream tasks.
Our extensive analysis shows that both ideas are essential for the success of CDLM, and work in synergy to set new state-of-the-art results for several multi-text tasks.\footnote{Code and models are available at \url{https://github.com/aviclu/CDLM}}
\end{abstract}
\section{Introduction}
\label{sec:intro}
The majority of NLP research addresses a \emph{single} text, typically at the sentence or document level. 
Yet, there are important applications which are concerned with aggregated information spread across multiple texts, such as cross-document coreference resolution~\cite{cybulska-vossen-2014-using}, classifying relations between document pairs~\cite{zhou-etal-2020-multilevel} and multi-hop question answering~\cite{yang-etal-2018-hotpotqa}.

Existing language models (LMs)~\cite{devlin2019bert,liu2019roberta,JMLR:v21:20-074}, which are pretrained with variants of the masked language modeling (MLM) self-supervised objective, are known to provide powerful representations for internal text structure~\cite{clark-etal-2019-bert,rogers-etal-2020-primer}, which were shown to be beneficial also for various multi-document tasks~\cite{Yang-Siamese-2020,zhou-etal-2020-multilevel}.

\begin{figure}
    \centering
    \footnotesize
    \frame{\includegraphics[width=1\linewidth]{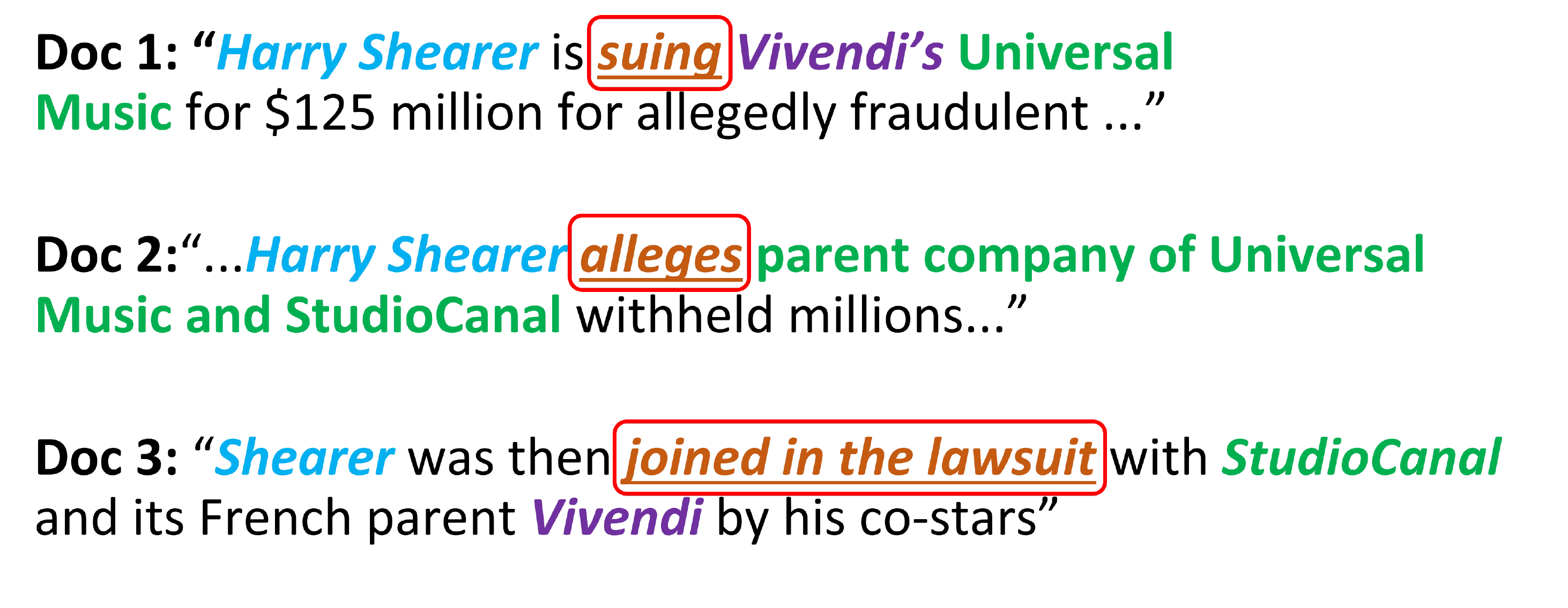}}
      \caption{An example from Multi-News~\cite{fabbri-etal-2019-multi}. Circled words represent matching events and the same color represents mention alignments.
    }
    \label{fig:comb_coref_examples}
    \vspace{-4mm}
\end{figure}

In this paper, we point out that beyond modeling internal text structure, multi-document tasks require also modeling cross-text relationships, particularly aligning or linking matching information elements across documents. For example, in Fig.~\ref{fig:comb_coref_examples}, one would expect a competent model to correctly capture that the two event mentions \textit{suing} and \textit{alleges}, from Documents 1 and 2, should be matched. 
Accordingly, capturing such cross-text relationships, in addition to representing internal text structure, can prove useful for downstream multi-text tasks, as we demonstrate empirically later. 

Following this intuition, we propose a new simple cross-document pretraining procedure, which is applied over sets of \emph{related} documents, in which informative cross-text relationships are abundant (e.g. like those in Fig.~\ref{fig:comb_coref_examples}). Under this setting, the model is encouraged to learn to consider and represent such relationships, since they provide useful signals when optimizing for the language modeling objective. 
For example, we may expect that it will be easier for a model to unmask the word \textit{alleges} in Document 2 if it would manage to effectively ``peek'' at Document 2, by matching the masked position and its context with the corresponding information in the other document. 

Naturally, considering cross-document context in pretraining, as well as in finetuning, requires a model that can process a fairly large amount of text. To that end, we leverage recent advances in developing efficient long-range transformers~\cite{beltagy2020longformer,zaheer2020big}, which utilize a global attention mode to build representations based on the entire input.
Overcoming certain restrictions in prior utilization of global attention (see Section \ref{sec:background}), we introduce a dynamic attention pattern during pretraining, over all masked tokens, and later utilize it selectively in finetuning.

Combining pretraining over related documents along with our global attention pattern yields a novel pretraining approach, that is geared to learn and implicitly encode informative cross-document relationships. As our experiments demonstrate, the resulting model, termed Cross-Document Language Model (CDLM), can be generically applied to downstream multi-document tasks, eliminating the need for task-specific architectures.
We show empirically that our model improves consistently over previous approaches in several tasks, including cross-document coreference resolution, multi-hop question answering, and document matching tasks. Moreover, we provide controlled experiments to ablate the two contributions of pretraining over related documents as well as new dynamic global attention. Finally, we provide additional analyses that shed light on the advantageous behavior of our CDLM. 
Our contributions are summarized below: 
\begin{itemize}[noitemsep,topsep=0pt]
    \item A new pretraining approach for multi-document tasks utilizing: (1) sets of related documents instead of single documents; (2) a new dynamic global attention pattern.
    \item The resulting model advances the state-of-the-art for several multi-document tasks.
\end{itemize}
\section{Method}
\subsection{Background: the Longformer Model}
\label{sec:background}

Recently, long-range LMs (e.g., Longformer~\cite{beltagy2020longformer}, BigBird~\cite{zaheer2020big}) have been proposed to extend the capabilities of earlier transformers \cite{vaswani2017attention} to process long sequences, using a sparse self-attention architecture. These models showed improved performance on both long-document and multi-document tasks~\cite{tay2021long}. In the case of multiple documents, instead of encoding documents separately, these models allow concatenating them into a long sequence of tokens and encoding them jointly. 
We base our model on Longformer, which sparsifies the full self-attention matrix in transformers by using a combination of a localized sliding window (called local attention), as well as a global attention pattern on a few specific input locations.
Separate weights are used for global and local attention. During pretraining, Longformer assigns \emph{local attention} to all tokens in a window around each token and optimizes the Masked Language Modeling (MLM) objective. Before task-specific finetuning, the attention mode is predetermined for each input token, assigning global attention to a few targeted tokens, such as special tokens, that are targeted to encode global information. 
Thus, in the Longformer model, global attention weights are not pretrained. Instead, they are initialized to the local attention values, before finetuning on each downstream task. We conjecture that the global attention mechanism can be useful for learning meaningful representations for modeling cross-document (CD) relationships. Accordingly, we propose augmenting the pretraining phase to exploit the global attention mode, rather than using it only for task-specific finetuning, as described below.

\subsection{Cross-Document Language Modeling}
\label{sec:cdmlm}

We propose a new pretraining approach consisting of two key ideas: (1) pretraining over sets of \emph{related} documents that contain overlapping information (2) pretraining with a dynamic global attention pattern over masked tokens, for referencing the entire cross-text context.

\paragraph{Pretraining Over Related Documents} Documents that describe the same topic, e.g., different news articles discussing the same story, usually contain overlapping information. Accordingly, various CD tasks may leverage from an LM infrastructure that encodes information regarding alignment and mapping across multiple texts. For example, for the case of CD coreference resolution, consider the underlined predicate examples in Figure~\ref{fig:comb_coref_examples}. One would expect a model to correctly align the mentions denoted by \textit{suing} and \textit{alleges}, effectively recognizing their cross-document relation.

Our approach to cross-document language modeling is based on pretraining the model on sets (clusters) of documents, all describing the same topic. 
Such document clusters are readily available in a variety of existing CD benchmarks, such as multi-document summarization (e.g., Multi-News~\cite{fabbri-etal-2019-multi}) and CD coreference resolution (e.g., ECB+~\cite{cybulska-vossen-2014-using}).
Pretraining the model over a set of related documents encourages the model to learn cross-text mapping and alignment capabilities, which can be leveraged for improved unmasking, as exemplified in Sec.~\ref{sec:intro}. 
Indeed, we show that this strategy directs the model to utilize information across documents and helps in multiple downstream CD tasks.

\begin{figure}[tb!!]
\centering
\includegraphics[width=1.\linewidth]{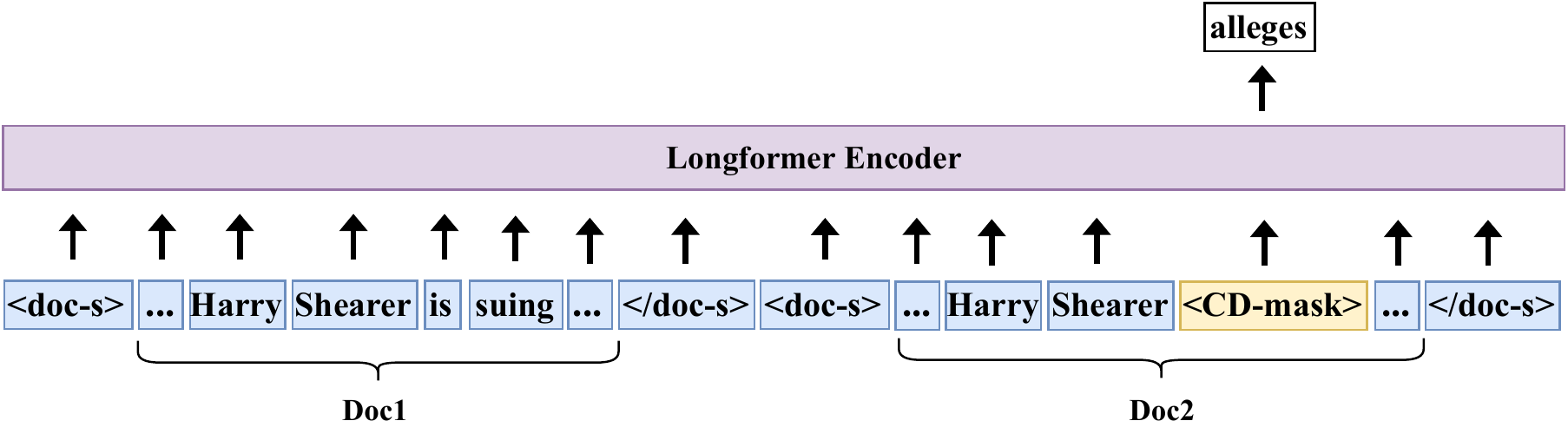}
\caption{CDLM pretraining: The input consists of concatenated documents, separated by special document separator tokens. The masked (unmasked) token colored in yellow (blue) represents global (local) attention. The goal is to predict the masked token \textit{alleges}, based on the global context, i.e, the entire set of documents.} \label{fig:cdmlm}
    \vspace{-5mm}
\end{figure}

\paragraph{Pretraining With Global Attention}
To support contextualizing information across multiple documents, we need to use efficient transformer models that scale linearly with input length. Thus, we base our cross-document language model (CDLM) on the Longformer model~\cite{beltagy2020longformer}, however, our setup is general and can be applied to other similar efficient Transformers. As described in Sec.~\ref{sec:background}, Longformer sparsifies the expensive attention operation for long inputs using a combination of local and global attention modes.
As input to the model, we simply concatenate related documents using new special document separator tokens, {\tt $\langle$doc-s$\rangle$} and {\tt $\langle$/doc-s$\rangle$}, for marking document boundaries. We apply a similar masking procedure as in BERT: For each training example, we randomly choose a sample of tokens (15\%) to be masked;\footnote{For details of masking see BERT \cite{devlin-etal-2019-bert}.} however, our pretraining strategy tries to predict each masked token while considering the \emph{full} document set, by assigning them \emph{global attention}, utilizing the global attention weights (see Section~\ref{sec:background}). This allows the Longformer to contextualize information both across documents as well as over long-range dependencies within-document. The non-masked tokens use local attention, by utilizing the local attention weights, as usual. 

An illustration of the CD pretraining procedure is depicted in Fig.~\ref{fig:cdmlm}, where the masked token associated with \textit{alleges} (colored in yellow) globally attends to the whole sequence, and the rest of the non-masked tokens (colored in blue) attend to their local context. With regard to the example in Fig.~\ref{fig:comb_coref_examples}, this masking approach aims to implicitly compel the model to learn to correctly predict the word \textit{alleges} by looking at the second document, optimally at the phrase \textit{suing}, and thus capture the alignment between these two events and their contexts.
\subsection{CDLM Implementation}
\label{sec:res}

In this section, we provide the experimental details used for pretraining our CDLM model. 

\paragraph{Corpus data} We use the preprocessed Multi-News dataset~\cite{fabbri-etal-2019-multi} as the source of related documents for pretraining. This dataset contains 44,972 training document clusters, originally intended for multi-document summarization. The number of source documents (that describe the same topic) per cluster varies from 2 to 10, as detailed in Appendix~\ref{subsec:multinews_data}. We consider each cluster of at least 3 documents for our cross-document pretraining procedure. We compiled our training corpus by concatenating related documents that were sampled randomly from each cluster, until reaching the Longformer's input sequence length limit of 4,096 tokens per sample. Note that this pretraining dataset is relatively small compared to conventional datasets used for pretraining. However, using it results in the powerful CDLM model.

\paragraph{Training and hyperparameters}
We pretrain the model according to our pretraining strategy, described in Section~\ref{sec:cdmlm}. We employ the Longformer-base model \cite{beltagy2020longformer} using the HuggingFace implementation \cite{wolf-etal-2020-transformers}
 and continue its pretraining, over our training data, for an additional 25k steps.\footnote{The training process for the base model takes 8 days on 8 RTX8000 GPUs. Training large models requires roughly 3x compute; therefore we do not focus on large models here and leave that for future work.}
 The new document separator tokens are added to the model vocabulary and randomly initialized before pretraining. 
 We use the same setting and hyperparameters as in \citet{beltagy2020longformer}, and as elaborated in Appendix~\ref{sec:hyper_cdlm}.
\section{Evaluations and Results}
\label{sec:eval}
This section presents experiments conducted to evaluate our \textsc{CDLM}, as well as the the ablations and baselines we used. For the intrinsic evaluation we measured the perplexity of the models. For extrinsic evaluations we considered event and entity cross-document coreference resolution, paper citation recommendation, document plagiarism detection, and multihop question answering. We also conducted an attention analysis, showing that our \textsc{CDLM} indeed captured cross-document and long-range relations during pretraining.\footnote{Since the underlying Longformer model is encoder-only, we evaluate on tasks that can be modeled using the encoder-only setting. We leave extensions to address seq2seq tasks like generation to future work.}

\paragraph{Baseline LMs} 
Recall that CDLM employs multiple related documents during pretraining, and assigns global attention to masked tokens. To systematically study the importance of these two components, we consider the following LM baselines:
\begin{itemize}[leftmargin=0pt,wide=0pt,label={--},topsep=0pt]
 \setlength\itemsep{-1pt}
\item  \textbf{Longformer}: the underlying Longformer model, without additional pretraining.
\item \textbf{Local \textsc{CDLM}}: pretrained using the same corpus of CDLM with the Longformer's attention pattern (local attention only). This baseline is intended to separate the effect of simply continuing pretraining Longformer on our new pre-training data. 
\item \textbf{Rand \textsc{CDLM}}: Longformer with the additional CDLM pretraining, while using random, unrelated documents from various clusters. This baseline model allows assessing whether pretraining using related documents is beneficial.
\item \textbf{Prefix \textsc{CDLM}}: pretrained similarly as CDLM but uses global attention for the first tokens in the input sequence, rather than the masked ones. This resembles the attention pattern of
\textsc{BigBird} \cite{zaheer2020big}, adopted for our cross-document setup. We use this ablation for examining this alternative global attention pattern, from prior work.
\end{itemize}

The data and pretraining hyperparameters used for the ablations above are the same as the ones used for our \textsc{CDLM} pretraining, except for the underlying Longformer, which is not further pretrained, and the Rand \textsc{CDLM}, that is fed with different document clusters (drawn from the same corpus). During all the experiments, the global attention weights used by the underlying Longformer and by Local \textsc{CDLM} are initialized to the values of their pretrained local attention weights. All the models above further finetune their global attention weights, depending on the downstream task.
When finetuning CDLM and the above models on downstream tasks involving multiple documents, we truncate the longer inputs to the Longformer's 4,096 token limit.

\subsection{Cross-Document Perplexity}
\label{subsec:cdppl}
First, we conduct a cross-document (CD) perplexity experiment, in a task-independent manner, to assess the contribution of the pretraining process. We used the Multi-News validation and test sets, each of them containing 5,622 document clusters, to construct the evaluation corpora. Then we followed the same protocol from the pretraining phase - 15\% of the input tokens are randomly masked, where the challenge is to predict the masked token given all documents in the input sequence. We matched the pretraining phase of each one of the ablation models: In \textsc{CDLM} and Rand \textsc{CDLM}, we assigned global attention for the masked tokens, and for Prefix \textsc{CDLM} the global attention is assigned to the 15\% first input tokens. Both Longformer and Local \textsc{CDLM} used local attention only. Perplexity is then measured by computing exponentiation of the loss.

The results are depicted in Table~\ref{tab:ppl}. The advantage of \textsc{CDLM} over Rand \textsc{CDLM}, which was pretrained equivalently over an equivalent amount of (unrelated) CD data, confirms that CD pretraining, over \textit{related} documents, indeed helps for CD masked token prediction across such documents.  Prefix \textsc{CDLM} introduces similar results since it was pretrained using a global attention pattern and the same corpora used by \textsc{CDLM}. The Local \textsc{CDLM} is expected to have difficulty to predict tokens across documents since it was pretrained without using global attention. Finally, the underlying Longformer model, which is reported as a reference point, is inferior to all the ablations since it was pretrained in a single document setting and without global attention or further pretraining on this domain.
Unlike the two local-attentive models, \textsc{CDLM} is encouraged to look at the full sequence when predicting a masked token. Therefore, as in the pretraining phase, it exploits related information in other documents, and not just the local context of the masked token, hence \textsc{CDLM}, as well as Prefix \textsc{CDLM}, result with a substantial performance gain.

\begin{table}[tb!]
\centering
    \footnotesize
\begin{tabular}{@{}lrr@{}}
                      \toprule
Model               & Validation           & Test     \\ \toprule
Longformer & 3.89 & 3.94\\
Local \textsc{CDLM} & 3.78 & 3.84 \\
Rand \textsc{CDLM} & 3.68 & 3.81\\
Prefix \textsc{CDLM} &\textbf{3.20} &3.41 \\
\textsc{CDLM} & 3.23 & \textbf{3.39}\\
\bottomrule
\end{tabular}
\caption{Cross-document perplexity evaluation on the validation and tests set of Multi-News. Lower is better.}
\label{tab:ppl}
\vspace{-5mm}
\end{table}

\subsection{Cross-Document Coreference Resolution}
\label{subsec:fdcdcoref}
Cross-document (CD) coreference resolution deals with identifying and clustering together textual mentions across multiple documents that refer to the same concept (see Fig.~\ref{fig:comb_coref_examples}). The considered mentions can be either entity mentions, usually noun phrases, or event mentions, typically verbs or nominalizations that appear in the text.

\paragraph{Benchmark.} We evaluated our CDLM by utilizing it over the ECB+ corpus~\cite{cybulska-vossen-2014-using}, the most commonly used dataset for CD coreference. ECB+ consists of within- and cross-document coreference annotations for entities and events (statistics are given in Appendix~\ref{subsec:ecb_data}). Following previous work, for comparison, we conduct our experiments on gold event and entity mentions.

\begin{table*}[!ht]
    \centering
    \resizebox{\textwidth}{!}{
    \begin{tabular}{@{}lllrrrrrrrrrrrrrrrrr@{}}
    \toprule
    &&& \multicolumn{3}{c}{MUC} && \multicolumn{3}{@{}c@{}}{$B^3$} & & \multicolumn{3}{c}{$CEAFe$} && \multicolumn{3}{c}{LEA} && CoNLL\\
    \cmidrule{4-6} \cmidrule{8-10} \cmidrule{12-14} \cmidrule{16-18} \cmidrule{20-20}
    &&& R & P & $F_1$ && R & P & $F_1$ && R &P & $F_1$ && R &P & $F_1$ && \multicolumn{1}{r}{$F_1$}  \\ 
   \midrule
   \multirow{6}{*}{\rotatebox[origin=c]{90}{Event}} 

        & \citet{barhom-etal-2019-revisiting} && 78.1 & 84.0 & 80.9 && 76.8 & 86.1 & 81.2 && 79.6 & 73.3 & 76.3 && 64.6 & 72.3 & 68.3 && 79.5\\
        
        & \citet{meged-etal-2020-paraphrasing} && 78.8 & 84.7 & 81.6 && 75.9 & 85.9 & 80.6 && 81.1 & 74.8 & 77.8 && 64.7 & 73.4 & 68.8 && 80.0\\

       & \citet{Cattan2020StreamliningCC} &&  85.1 & 81.9 & 83.5 && 82.1 & 82.7 & 82.4 && 75.2 & 78.9 & 77.0 && 68.8 & 72.0 & 70.4 && 81.0 \\
        
     &\citet{zeng-etal-2020-event} &&  85.6 & 89.3 & 87.5 && 77.6 & 89.7 & 83.2 && 84.5 & 80.1 & \textbf{82.3}  && - & - & - &&  84.3   \\
        
      &   \citet{yu2020paired} &&  88.1 & 85.1 & 86.6 && 86.1 & 84.7 & 85.4 && 79.6 & 83.1 & 81.3 && - & - & - &&  84.4   \\
        
         &\citet{allaway2021sequential} &&  81.7 & 82.8 & 82.2 && 80.8 & 81.5 & 81.1 && 79.8  & 78.4 & 79.1  && - & - & - &&  80.8 \\

        & \textsc{CDLM} && 87.1  & 89.2 & \textbf{88.1} && 84.9 & 87.9 & \textbf{86.4} && 83.3 & 81.2 & 82.2 && 76.7 & 77.2 & \textbf{76.9} && \textbf{85.6} \\

    \midrule
    \multirow{3}{*}{\rotatebox[origin=c]{90}{Entity}}  

     &\citet{barhom-etal-2019-revisiting} &&  81.0 & 80.8 & 80.9 && 66.8 & 75.5 & 70.9 && 62.5 & 62.8 & 62.7 && 53.5 & 63.8 & 58.2 && 71.5\\
      &  \citet{Cattan2020StreamliningCC} && 85.7 & 81.7 & 83.6 && 70.7 & 74.8 & 72.7 && 59.3 & 67.4 & 63.1 && 56.8 & 65.8 & 61.0 && 73.1\\
     &\citet{allaway2021sequential} &&   83.9 & 84.7 & 84.3 && 74.5 & 70.5 & 72.4 && 70.0  & 68.1 & 69.2  && - & - & - &&  75.3   \\

        & \textsc{CDLM} && 88.1  & 91.8 &\textbf{89.9} && 82.5 & 81.7 & \textbf{82.1} && 81.2 & 72.9 & \textbf{76.8} && 76.4 & 73.0 & \textbf{74.7} && \textbf{82.9} \\
   
    \bottomrule
    \end{tabular}}
    \caption{Results on event and entity cross-document coreference resolution on ECB+ test set.}
    \label{tab:subtopic_results_event}
    \vspace{-3mm}
\end{table*}

We follow the standard coreference resolution evaluation metrics: \textit{MUC}~\cite{vilain1995model}, $\textit{B}^\textbf{3}$~\cite{bagga1998algorithms}, \textit{CEAFe}~\cite{luo2005coreference}, their average \textit{CoNLL F1}, and the more recent \textit{LEA} metric~\cite{moosavi-strube-2016-coreference}.

\paragraph{Algorithm.} 
Recent approaches for CD coreference resolution train a pairwise scorer to learn the probability that two mentions are co-referring. At inference time, an agglomerative clustering based on the pairwise scores is applied, to form the coreference clusters. We made several modifications to the pairwise scorer.
The current state-of-the-art models~\cite{zeng-etal-2020-event,yu2020paired} train the pairwise scorer by including only the local contexts (containing sentences) of the candidate mentions. They concatenate the two input sentences and feed them into a transformer-based LM. Then, part of the resulting tokens representations are aggregated into a single feature vector which is passed into an additional MLP-based scorer to produce the coreference probability estimate.
To accommodate our proposed CDLM model, we modify this modeling by including the entire documents containing the two candidate mentions, instead of just their containing sentences, and assigning the global attention mode to the mentions' tokens and to the {\tt [CLS]} token. The full method and hyperparameters are elaborated in Appendix~\ref{subsec:ft_coref}.

\paragraph{Baselines.} We consider state-of-the-art baselines that reported results over the ECB+ benchmark. The following baselines were used for both event and entity coreference resolution:
\begin{itemize}[leftmargin=0pt,wide=0pt,label={--},topsep=0pt]
 \setlength\itemsep{-1pt}
\item \citet{barhom-etal-2019-revisiting} is a model trained jointly for solving event and entity coreference as a single task. It utilizes semantic role information between the candidate mentions.
\item \citet{Cattan2020StreamliningCC} is a model trained in an end-to-end manner (jointly learning mention detection and coreference following~\citet{lee-etal-2017-end}), employing the RoBERTa-large model to encode each document separately and to train a pair-wise scorer atop.
\item \citet{allaway2021sequential} is a BERT-based model combining sequential prediction with incremental clustering.
\end{itemize}

\begin{table}[!tb]
    \newcommand{\colindent}{\;}
    \centering
    \resizebox{0.38\textwidth}{!}{
    \footnotesize
    \begin{tabular}{@{}lrr@{}}
    \toprule
    \phantom{fwidsvhckzxjvchndfzxvvgdaczfc} & F1 & $\Delta$\\
    \midrule
    full document \textsc{CDLM}  & 85.6 & \\
    \colindent $-$ sentences only \textsc{CDLM} & 84.2 & -1.4 \\
    \colindent $-$  Longformer & 84.6 & -1.0\\
    \colindent $-$  Local \textsc{CDLM} & 84.7 & -0.9 \\
    \colindent $-$  Rand \textsc{CDLM} & 84.1 & -1.5 \\
    \colindent $-$  Prefix \textsc{CDLM} & 85.1 & -0.5 \\
    \bottomrule
    \end{tabular}}
    \caption{Ablation results (CoNLL F1) on our model on the test set of ECB+ event coreference.}
    \label{tab:coref_ablations}
    \vspace{-3mm}
\end{table}

The following baselines were used for event coreference resolution. They all integrate external linguistic information as additional features.
\begin{itemize}[leftmargin=0pt,wide=0pt,label={--},topsep=0pt]
 \setlength\itemsep{-1pt}
\item \citet{meged-etal-2020-paraphrasing} is an extension of \citet{barhom-etal-2019-revisiting}, leveraging external knowledge acquired from a paraphrase resource~\cite{shwartz-etal-2017-acquiring}.
\item \citet{zeng-etal-2020-event} is an end-to-end model, encoding the concatenated two sentences containing the two mentions by the BERT-large model. Similarly to our algorithm, they feed a MLP-based pairwise scorer with the concatenation of the {\tt [CLS]} representation and an attentive function of the candidate mentions representations.
\item \citet{yu2020paired} is an end-to-end model similar to \citet{zeng-etal-2020-event}, but uses rather RoBERTa-large and does not consider the {\tt [CLS]} contextualized token representation for the pairwise classification. %
\end{itemize}

\paragraph{Results.} The results on event and entity CD coreference resolution are depicted in Table \ref{tab:subtopic_results_event}.
Our \textsc{CDLM} outperforms all methods, including the recent sentence based models on event coreference. All the results are statistically significant using bootstrap and permutation tests with ${p<0.001}$~\cite{dror-etal-2018-hitchhikers}. \textsc{CDLM} largely surpasses state-of-the-art results on entity coreference, even though these models utilize external information and use large pretrained models, unlike our base model. In Table~\ref{tab:coref_ablations}, we provide the ablation study results. Using our model with sentences only, i.e., considering only the sentences where the candidate mentions appear (as the prior baselines did), exhibits lower performance, resembling the best performing baselines. Some crucial information about mentions can appear in a variety of locations in the document, and is not concentrated in one sentence. This characterizes long documents, where pieces of information are often spread out.
Overall, the ablation study shows the advantage of using our pretraining method, over related documents and using a scattered global attention pattern, compared to the other examined settings. 
Recently, our CDLM-based coreference model was utilized to generate event clusters within an effective faceted-summarization system for multi-document exploration \cite{iFacetSum}.
\subsection{Document matching}
\label{subsec:aan}

We evaluate our CDLM over document matching tasks, aiming to assess how well our model can capture interactions across multiple documents. We use the recent multi-document classification benchmark by \citet{zhou-etal-2020-multilevel} which includes two tasks of citation recommendation and plagiarism detection. The goal of both tasks is categorizing whether a particular relationship holds between two input documents.
Citation recommendation deals with detecting whether one reference document should cite the other one, while the plagiarism detection task infers whether one document plagiarizes the other one. To compare with recent state-of-the-art models, we utilized the setup and data selection from~\citet{zhou-etal-2020-multilevel}, which provides three datasets for citation recommendation and one for plagiarism detection. 

\paragraph{Benchmarks.} For citation recommendation, the datasets include the ACL Anthology Network Corpus \citep[AAN;][]{radev2013acl}, the Semantic Scholar Open Corpus \citep[OC;][]{bhagavatula-etal-2018-content}, and the Semantic Scholar Open Research Corpus \citep[S2ORC;][]{lo-etal-2020-s2orc}. For plagiarism detection, the dataset is the Plagiarism Detection Challenge \citep[PAN;][]{potthast2013overview}. 

AAN is composed of computational linguistics papers which were published on the ACL Anthology from 2001 to 2014, OC is composed of computer science and neuroscience papers, S2ORC is composed of open access papers across broad domains of science, and PAN is composed of web documents that contain several kinds of plagiarism phenomena. For further dataset prepossessing details and statistics, see Appendix~\ref{subsec:cda_data}.

\paragraph{Algorithm.} For our models, we added the {\tt [CLS]} token at the beginning of the input sequence, assigned it global attention, and concatenated the pair of texts, according to the finetuning setup discussed in Section~\ref{sec:cdmlm}. The hyperparameters are further detailed in Appendix~\ref{subsec:app_aan}.

\paragraph{Baselines.} We consider the reported results of the following recent baselines:
\begin{itemize}[leftmargin=0pt,wide=0pt,label={--},topsep=0pt]
 \setlength\itemsep{-1pt}
\item \textbf{\textsc{HAN}} \cite{yang-etal-2016-hierarchical} proposed the Hierarchical Attention Networks (HANs). These models employ a bottom-up approach in which a document is represented as an aggregation of smaller components i.e., sentences, and words. They set competitive performance in different tasks involving long document encoding \cite{sun-etal-2018-stance}. 
\item \textbf{\textsc{SMASH}} \cite{jiang2019semantic} is an attentive hierarchical recurrent neural network (RNN) model, used for tasks related to long documents. 
\item \textbf{\textsc{SMITH}} \cite{Yang-Siamese-2020} is a BERT-based hierarchical model, similar HANs.
\item \textbf{\textsc{CDA}} \cite{zhou-etal-2020-multilevel} is a cross-document attentive mechanism (CDA) built on top of HANs, based on BERT or GRU models (see Section~\ref{sec:related}). %

\end{itemize}

Both \textsc{SMASH} and \textsc{SMITH} reported results only over the AAN benchmark. In addition, they used a slightly different version of the AAN dataset,\textsuperscript{\ref{note1}} and included the full documents, unlike the dataset that \cite{zhou-etal-2020-multilevel} used, which we utilized as well, that considers only the documents' abstracts.  

\begin{table}
        \centering
        \footnotesize
        \def\arraystretch{1.0}\tabcolsep=4pt    
        \begin{tabular}[b]{@{}lllll@{}}
          \toprule
            Model             & AAN              & OC           & S2orc           & PAN  \\ \toprule
              \textsc{SMASH} \citeyearpar{jiang2019semantic}\footnotemark\label{note1}         & 80.8 &   - & -  &  - \\ 
              \textsc{SMITH} \citeyearpar{Yang-Siamese-2020}\textsuperscript{\ref{note1}}       & 85.4 &   -    & - &   - \\ 
            \textsc{BERT-HAN} \citeyearpar{zhou-etal-2020-multilevel}    & 65.0 &    86.3      &   90.8    &    \textbf{87.4} \\             
            \textsc{GRU-HAN+CDA} \citeyearpar{zhou-etal-2020-multilevel}    & 75.1 &    89.9      &   91.6    &    78.2 \\   
            \textsc{BERT-HAN+CDA} \citeyearpar{zhou-etal-2020-multilevel}    & 82.1 &    87.8      &   92.1    &    86.2 \\             
              \midrule
             Longformer    &85.4&         93.4         &    95.8   &  80.4  \\ 
              Local \textsc{CDLM}    & 83.8 &    92.1         &    94.5   & 80.9  \\ 
               Rand \textsc{CDLM}     & 85.7&   93.5      &   94.6     &    79.4  \\ 
             Prefix \textsc{CDLM}   & 87.3  &   94.8  &  94.7     &  81.7  \\ 
             \textsc{CDLM}      & \textbf{88.8}   &          \textbf{95.3}     &  \textbf{96.5}  & 82.9\\
            \bottomrule
        \end{tabular}
        \caption{$F_1$ scores over the document matching benchmarks' test sets.}
        \label{tab:aan2}
        \vspace{-5mm}
\end{table}

\paragraph{Results.} The results on the citation recommendation and plagiarism detection tasks are depicted in Table~\ref{tab:aan2}. We observe that even though \textsc{SMASH} and \textsc{SMITH} reported results using the full documents for the AAN task, our model outperforms them, using the partial version of the dataset, as in~\citet{zhou-etal-2020-multilevel}. Moreover, unlike our model, CDA is task-specific since it trains new cross-document weights for each task, yet it is still inferior to our model, evaluating on the three citation recommendation benchmarks. On the plagiarism detection benchmark, interestingly, our models does not perform better. Moreover, CDA impairs the performance of \textsc{BERT-HAN}, implying that dataset does not require detailed cross-document attention at all. In our experiments, finetuning \textsc{BERT-HAN+CDA} over the PAN dataset yielded poor results: $F_1$ score of 79.6, substantially lower compared to our models. The relatively small size of PAN may explain such degradations. 

\footnotetext{Following the most recent work of \citet{zhou-etal-2020-multilevel}, we evaluate our model on their version of the dataset. We also quote the results of \textsc{SMASH} and \textsc{SMITH} methods, even though they used a somewhat different version of this dataset, hence their results are not fully comparable to the results of our model and those of \textsc{CDA}.}

\subsection{Multihop Question answering}
\label{subsec:hpqa}
In the task of multihop question answering, a model is queried to extract answer spans and evidence sentences, given a question and multiple paragraphs from various related and non-related documents. This task includes challenging questions, that answering them requires finding and reasoning over multiple supporting documents.

\paragraph{Benchmark.} We used the HotpotQA-distractor dataset~\cite{yang-etal-2018-hotpotqa}. Each example in the dataset is comprised of a question and 10 different paragraphs from different documents, extracted from Wikipedia; two gold paragraphs include the relevant information for properly answering the question, mixed and shuffled with eight distractor paragraphs (for the full dataset statistics, see~\citet{yang-etal-2018-hotpotqa}). There are two goals for this task: extraction of the correct answer span, and detecting the supporting facts, i.e., evidence sentences. 

\paragraph{Algorithm.} We employ the exact same setup from~\cite{beltagy2020longformer}: We concatenate all the 10 paragraphs into one large sequence, separated by document separator tokens, and using special sentence tokens to separate sentences. The model is trained jointly in a multi-task manner, where classification heads specialize on each sub-task, including relevant paragraphs prediction, evidence sentences identification, extracting answer spans and inferring the question types (yes, no, or span). For details and hyperparameters, see Appendix~\ref{subsec:multihopqa} and ~\citet[Appendix D]{beltagy2020longformer}.

\begin{table}[t]
\centering
\footnotesize
\setlength{\tabcolsep}{4pt} 
        \begin{tabular}[b]{@{}lrrr@{}}
            \toprule
            Model            & Ans & Sup & Joint \\ \midrule
            Transformer-XH \citeyearpar{Zhao2020Transformer} & 66.2 &  72.1 & 52.9\\
            Graph Recurrent
Retriever \citeyearpar{Asai2020Learning} & 73.3 & 76.1 & 61.4\\
            RoBERTa-lf   \citeyearpar{beltagy2020longformer}  & 73.5   & 83.4    & 63.5     \\
            \textsc{BigBird}  \citeyearpar{zaheer2020big} & \textbf{75.5}&  \textbf{87.1}& \textbf{67.8}\\
              \midrule
            Longformer  & 74.5   & 83.9    & 64.5     \\ 
            Local \textsc{CDLM} & 74.1 & 84.0 & 64.2 \\
            Rand \textsc{CDLM} & 72.7 & 84.8 &  63.7 \\
            Prefix \textsc{CDLM} & 74.8 & 84.7 & 65.2\\
            \textsc{CDLM} & 74.7 & 86.3 & 66.3\\
            \bottomrule
        \end{tabular}
\caption{HotpotQA-distractor results ($F_1$) for the dev set. We use the ``base'' model size results from prior work for direct comparison. Ans: answer span, Sup: Supporting facts.} 
\label{tab:hotpotqa}
\vspace{-5mm}
\end{table}

\paragraph{Results.} The results are depicted in Table~\ref{tab:hotpotqa}, where we included also the results for Transformer-XH~\cite{Zhao2020Transformer}, a transformer-based model that constructs global contextualized representations, Graph Recurrent Retriever~\cite{Asai2020Learning}, a recent strong graph-based passage retrieval method, RoBERTa~\cite{liu2019roberta}, which was modified by \citet{beltagy2020longformer} to operate on long sequences (dubbed RoBERTa-lf), and \textsc{BigBird}~\cite{zaheer2020big}, a long-range transformer model which was pretrained on a massive amount of text.
\textsc{CDLM} outperforms all the ablated models as well as the comparably sized models from prior work (except for \textsc{BigBird}), especially in the supporting evidence detection sub-task. We note that the \textsc{BigBird} model was pretrained on much larger data, using more compute resources compared both to the Longformer model and to our models. We suspect that with more compute and data, it is possible to close the gap between CDLM and \textsc{BigBird} performance. We leave for future work evaluating a larger version of the \textsc{CDLM} model against large, state-of-the-art models.
\subsection{Attention Analysis}
\label{subsec:qualitative}
It was recently shown that during the pretraining phase, LMs learn to encode various types of linguistic information, that can be identified via their attention patterns~\cite{wiegreffe-pinter-2019-attention,rogers2020primer}. In \citet{clark-etal-2019-bert}, the attention weights of BERT were proved as informative for probing the degree to which a particular token is ``important'', as well as its linguistic roles. For example, they showed that the averaged attention weights from the last layer of BERT are beneficial features for dependency parsing.

We posit that our pretraining scheme, which combines global attention and a multi-document context, captures alignment and mapping information across documents. Hence, we hypothesize that the global attention mechanism favors cross-document (CD), long-range relations. To gain more insight, our goal is to investigate if our proposed pretraining method leads to relatively higher global attention weights between co-referring mentions compared to non-co-referring ones, even without any finetuning over CD coreference resolution.

\paragraph{Benchmark.} We randomly sampled 1,000 positive and 1,000 negative coreference-pair examples from the ECB+ CD coreference resolution benchmark, for both events and entities. Each example consists of two concatenated documents and two coreference candidate mentions (see Section~\ref{subsec:fdcdcoref}).

\paragraph{Analysis Method.} For each example, which contains two mention spans, we randomly pick one to be considered as the \emph{source span}, while the second one is the \emph{target span}. We denote the set of the tokens in the source and target spans as $S$ and $T$, respectively. Our goal is to quantify the degree of alignment between $S$ and $T$, using the attention pattern of the model. We first assign global attention to the tokens in the source span (in $S$). Next, we pass the full input through the model, compute the normalized attention weights for all the tokens in the input with respect to $S$, by aggregating the scores extracted from the last layer of the model. The score for an input token $i\notin S$, is given by
\[
s(i|S) \propto \exp \left[\sum_{k=1}^n \sum_{j\in S} \left(\alpha_{i,j}^k+\alpha_{j,i}^k\right)\right],
\]
where ${\alpha_{i,j}^k}$ is the \emph{global} attention weight from token $i$ to token $j$ produced by head $k$, and $n$ is the total number of attention heads (the score is computed using only the last layer of the model). Note that we include both directions of attention. The target span score is then given by ${s(T|S)=\frac{1}{|T|}\sum_{j\in T}s(j|S)}$. Finally, we calculate the percentile rank (PR) of ${s(T|S)}$, compared to the rest of the token scores within the containing document of $T$, namely, ${\{s(i|S)|i\notin T\}}$. %

For positive coreference examples, plausible results are expected to be associated with high attention weights between the source and the target spans, resulting with a high value of ${s(T|S)}$, and thus, yielding a higher PR. For negative examples, the target span is not expected to be promoted with respect to the rest of the tokens in the document.

\begin{figure}
\centering
\footnotesize
\begin{tabular}{@{}p{\linewidth}@{}}
\toprule
\textbf{Doc 1}:
President Obama will \textbf{\underline{\textcolor{blue}{name}}} Dr. Regina Benjamin as U.S. Surgeon General in a Rose Garden announcement late this morning. Benjamin, an Alabama family physician, [...]\\

\textbf{Doc 2}:
[...] Obama \textbf{\underline{\textcolor{darkspringgreen}{nominates}}} new surgeon general: MacArthur \lq\lq genius grant \rq\rq fellow Regina Benjamin. [...] \\
\bottomrule
\end{tabular}
\caption{An example from ECB+ corpus. The underlined phrases represent a positive, co-referring event mention pair. The blue (green) colored mention is considered as the \textit{source (target)} span.}
\label{fig:ecb_example}
\vspace{-5mm}
\end{figure}

\paragraph{Results.} First, we apply the procedure above over one selected example, depicted in Figure~\ref{fig:ecb_example}.  We consider the two CD co-referring event mentions: \textit{name} and \textit{nominates} as the source and target spans, respectively. The target span received a PR of 69\% when evaluating the underlying Longformer. Notably, it received a high PR of 90\% when using our \CDLM, demonstrating the advantage of our novel pretraining method. Next, we turn to a systematic experiment, elucidating the relative advantage of pretraining with global attention across related documents. In Table~\ref{tab:pr_ecb}, we depict the mean PR (MPR) computed over all the sampled examples, for all our pretrained models. We observe that none of the models fail\footnote{Typically, PR of $\sim$50\% corresponds to random ranking.} on the set of negatives, since the negative examples contain reasonable event or entity mentions, rather than random, non informative spans. For the positive examples, the gap of up to 10\% of MPR between the ``Local'' and ``Global'' models shows the advantage of adopting global attention during the pretraining phase. This indicates that the global attention mechanism implicitly helps to encode alignment information.

\begin{table}[!t]
\renewcommand{\arraystretch}{1.1}
\setlength{\tabcolsep}{5pt}
\centering
    \footnotesize
    \begin{tabular}{@{}llccccccc@{}}
    \toprule
    && \multicolumn{2}{c}{Pos. MPR (\%)} && \multicolumn{2}{c}{Neg. MPR (\%)} \\
    \cmidrule{3-4} \cmidrule{6-7} 
    && events & entities  && events & entities  \\ 
  \midrule
    \multirow{2}{*}{\rotatebox[origin=c]{90}{Local}} 
    &Longformer & 61.9 & 59.7 && 54.8 & 50.5\\
    &Local \textsc{CDLM} & \textbf{62.2} & \textbf{60.8} && 54.6 & 52.6 \\
    \midrule
    \multirow{3}{*}{\rotatebox[origin=c]{90}{Global}} 
    &Rand \textsc{CDLM} & 70.6 & 69.1 && 56.6& 53.2\\
    &Prefix \textsc{CDLM} & 70.7 & 69.4 && 58.5 & 56.5\\
    &\textsc{CDLM} & \textbf{72.1} & \textbf{70.3} && 58.0 & 55.7\\
    \bottomrule
    \end{tabular}
    \caption{Cross-document coreference resolution alignment MPR scores of the target span, with respect to the tokens in the same document.}
    \label{tab:pr_ecb}
    \vspace{-5mm}
\end{table}

\section{Related Work}
\label{sec:related}

Recently, long-context language models \cite{beltagy2020longformer,zaheer2020big} introduced the idea of processing multi-document tasks using a single long-context sequence encoder.
However, pretraining objectives in these models consider only single documents. Here, we showed that additional gains can be obtained by MLM pretraining using multiple \textit{related} documents as well as a new dynamic global attention pattern.

Processing and aggregating information from multiple documents has been also explored in the context of document retieval, aiming to extract information from a large set of documents~\cite{guu2020realm,lewis2020pre,patrick2020retrieval,karpukhin-etal-2020-dense}. These works focus on retrieving relevant information from often a large collection of documents, by utilizing short-context LMs, and then generate information of interest. CDLM instead provides an approach for improving the encoding and contextualizing information across multiple documents. As opposed to the mentioned works, our model utilizes long-context LM and can include broader contexts of more than a single document.

The use of cross-document attention has been recently explored by the Cross-Document Attention (CDA) \cite{zhou-etal-2020-multilevel}. CDA specifically encodes two documents, using hierarchical attention networks, with the addition of cross attention between documents, and makes similarity decision between them. Similarly, the recent DCS model~\cite{ginzburg-etal-2021-self} suggested a cross-document finetuning scheme for unsupervised document-pair matching method (processing only two documents at once). Our CDLM, by contrast, is a general pretrained language model that can be applied to a variety of multi-document downstream tasks, without restrictions on the number of input documents, as long as they fit the input length of the Longformer.

Finally, our pretraining scheme is conceptually related to cross-encoder models that leverage simultaneously multiple related information sources. For example, the Translation Language Model (TLM) \cite{conneau2019cross} encodes together sentences and their translation, while certain cross-modality encoders pretrain over images and texts in tandem (e.g., ViLBERT~\cite{lu2019vilbert}).
\section{Conclusion}
We presented a novel pretraining strategy and technique for cross-document language modeling, providing better encoding for cross-document (CD) downstream tasks. 
Our contributions include the idea of leveraging clusters of related documents for pretraining, via cross-document masking, along with a new long-range attention pattern, together driving the model to learn to encode CD relationships. This was achieved by extending the global attention mechanism of the Longformer model to apply already in pretraining, creating encodings that attend to long-range information across and within documents.
Our experiments assess that our cross-document language model yields new state-of-the-art results over several CD benchmarks, while, in fact, employing substantially smaller models. Our analysis showed that CDLM implicitly learns to recover long-distance CD relations via the attention mechanism. We propose future research to extend this framework to train larger models, and to develop cross-document sequence-to-sequence models, which would support CD tasks that involve a generation phase.
\section*{Acknowledgments}
We thank Doug Downey and Luke Zettlemoyer for fruitful discussions and helpful feedback, and Yoav Goldberg for helping us connect with collaborators on this project. The work described herein was supported in part by grants from Intel Labs, the Israel Science Foundation grant 1951/17, the Israeli Ministry of Science and Technology, and the NSF Grant OIA-2033558.

\bibliography{anthology,emnlp2021}
\bibliographystyle{acl_natbib}

\clearpage
\appendix
\section{Dataset Statistics and Details}

In this section, we provide details regrading the pretraining corpus and benchmarks we used during our experiments.

\subsection{Multi-News Corpus}
\label{subsec:multinews_data}
We used the preprocessed, not truncated version of Multi-News, which totals 322MB of uncompressed text.\footnote{We used the dataset available in \url{https://drive.google.com/open?id=1qZ3zJBv0zrUy4HVWxnx33IsrHGimXLPy}.} Each one of the preprocessed documents contains up to 500 tokens. The average and 90$^\text{th}$ percentile of input length is 2.5k and 3.8K tokens, respectively. In Table~\ref{tab:multi_stats} we list the number of related documents per cluster. This follows the original dataset construction suggested in~\citet{fabbri-etal-2019-multi}. 
\begin{table}[htp]
\centering
    \small
  \def\arraystretch{1.12}\tabcolsep=7pt    
\begin{tabular}{ll}
                      \toprule
\# of docs in cluster               &  Frequency            \\ \toprule
 3 & 12,707      \\
 4 & 5,022     \\
 5 &  1,873      \\
 6 & 763     \\
 7 & 382     \\
 8 & 209     \\
 9 & 89     \\
10 & 33     \\
\midrule
\textbf{Total}  & 21,078     \\

\bottomrule
\end{tabular}
\caption{MultiNews training set statistics.}\smallskip 
\label{tab:multi_stats}
\end{table}

\subsection{ECB+ Dataset}
\label{subsec:ecb_data}
In Table~\ref{tab:ecb_stat}, we list the statistics about training, development, and test splits regarding the topics, documents, mentions and coreference clusters. We follow the data split used by previous works~\cite{cybulska2015bag, kenyon-dean-etal-2018-resolving,barhom-etal-2019-revisiting}: For training, we consider the topics: 1, 3, 4, 6-11, 13- 17, 19-20, 22, 24-33; For Validation, we consider the topics: 2, 5, 12, 18, 21, 23, 34, 35; For test, we consider the topics: 36-45.
\begin{table}[!hbt]
    \centering
        \footnotesize
    \begin{tabular}{@{}lccc@{}}
    \toprule
    & Train & Validation & Test \\
    \midrule
    Topics & 25 & 8 & 10 \\
    Docs & 594 & 196 & 206 \\
    Mentions & 3808/4758 & 1245/1476 & 1780/2055 \\
    Clusters & 411/472 & 129/125 & 182/196 \\
    \bottomrule
    \end{tabular}
    \caption{ECB+ dataset statistics. The slash numbers for Mentions and Clusters represent event/entity statistics.}
    \label{tab:ecb_stat}
\end{table}

\subsection{Paper Citation Recommendation \& Plagiarism Detection Datasets}
\label{subsec:cda_data}
In Table~\ref{tab:splits}, we list the statistics about training, development, and test splits for each benchmark separatly, and in Table~\ref{tab:counts}, we list the document and example counts for each benchmark. The statistics are taken from~\citet{zhou-etal-2020-multilevel}.
\begin{table}[htp]
\centering
    \small
  \def\arraystretch{1.2}\tabcolsep=7.5pt    
\begin{tabular}{lccc}
\toprule
  Dataset  & Train& Validation & Test \\
  \toprule
AAN         & 106,592 & 13,324   & 13,324  \\ 
OC          & 240,000 & 30,000  & 30,000 \\ 
S2ORC       & 152,000   & 19000   & 19000  \\
PAN         & 17,968 & 2,908 & 2,906 \\
\toprule
\end{tabular}
\caption{Document-to-Document benchmarks statistics: Details regrading the training, validation, and test splits.} 
\label{tab:splits}
\end{table}

\begin{table}[htp]
      \centering
          \small
  \def\arraystretch{1.2}\tabcolsep=12pt
\begin{tabular}{lcc}
\toprule
    Dataset& \# of doc pairs& \# of docs \\  \toprule
AAN         & 132K & 13K     \\ 
OC & 300K   & 567K    \\ 
S2ORC       & 190K    & 270K    \\
PAN         & 34K & 23K   \\
\toprule
\end{tabular}
\caption{Document-to-Document benchmarks statistics: The reported numbers are the count of document pairs and the count of unique documents.}
\label{tab:counts}
\end{table}

The preprocessing of the datasets performed by~\citet{zhou-etal-2020-multilevel} includes the following steps: For AAN, only pairs of documents that include abstracts are considered, and only their abstracts are used. For OC, only one citation per paper is considered, and the dataset was downsampled significantly. For S2ORC, formed pairs of citing sections and the corresponding abstract in the cited paper are included, and the dataset was downsampled significantly. For PAN, pairs of relevant segments out of the entire document were extracted.

For all the datasets, negative pairs were sampled randomly. Then, a standard preprocessing that includes filtering out characters that are not digits, letters, punctuation, or white space in the texts was performed.

\begin{figure*}[tb]
\centering
\includegraphics[width=0.79\linewidth]{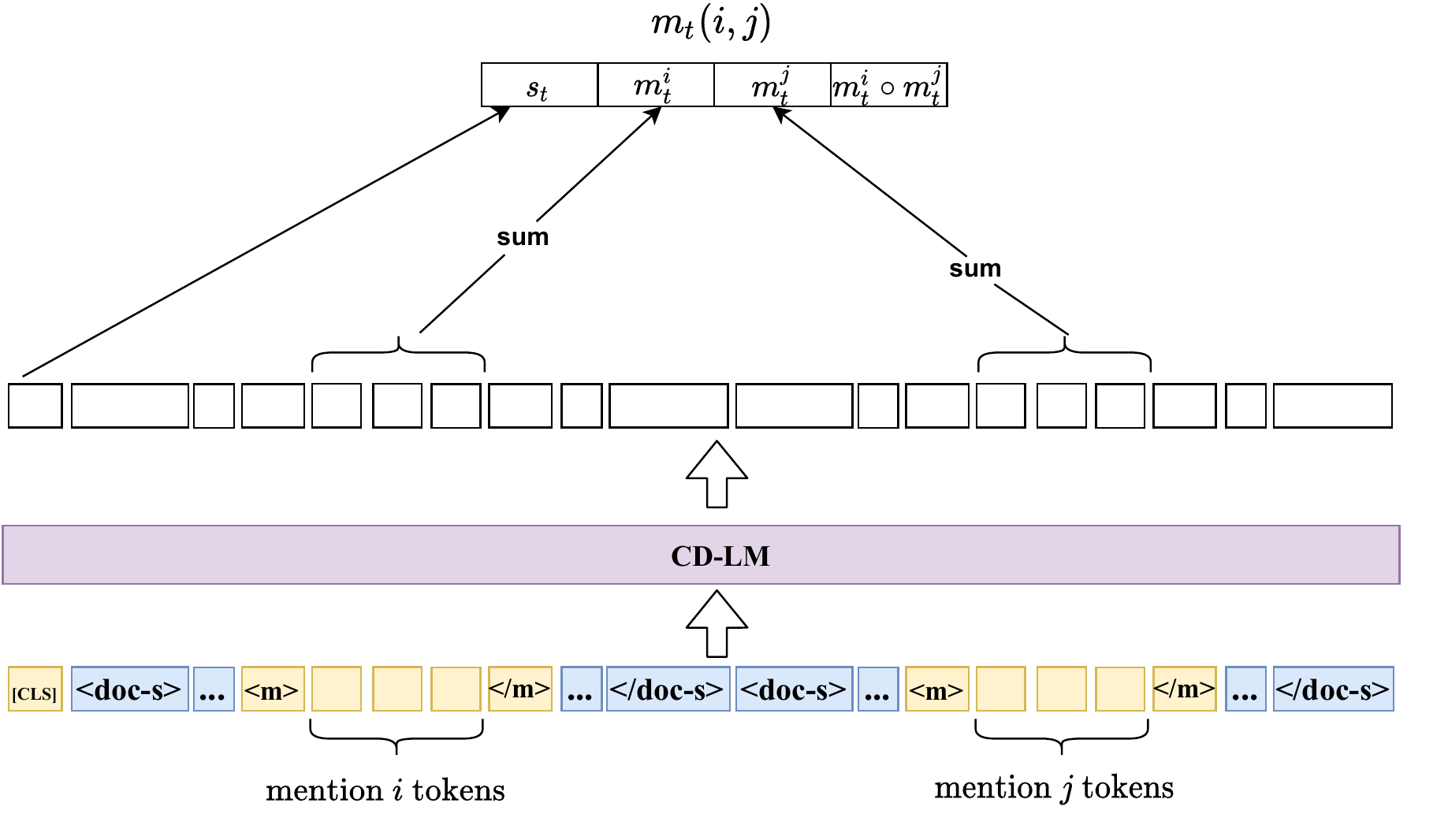}
\caption{CD-coreference resolution pairwise mention representation, using the new setup, for our \textsc{CDLM} models. $m_t^i$, $m_t^j$ and $s_t$ are the cross-document contextualized representation vectors for mentions $i$ and $j$, and of the {\tt [CLS]} token, respectively. ${m_t^i \circ m_t^j}$ is the element-wise product between $m_t^i$ and $m_t^j$. $m_t(i,j)$ is the final produced pairwise-mention representation. The tokens colored in yellow represent global attention, and tokens colored in blue represent local attention.} 
\label{fig:cdcoref}
\end{figure*}

\section{CDLM Pretraining Hyperparameters}
\label{sec:hyper_cdlm}
In this section, we detail the hyperparameters setting of the models we pretrained, including \textsc{CDLM} Prefix \textsc{CDLM}, Rand \textsc{CDLM}, and Local \textsc{CDLM}: The input sequences are of the length of 4,096, effective batch size of 64 (using gradient accumulation and batch size of 8), a maximum learning rate of 3e-5, and a linear warmup of 500 steps, followed by a power 3 polynomial decay. For speeding up the training and reducing memory consumption, we used the mixed-precision (16-bits) training mode. The pretraining took 8 days, using eight 48GB RTX8000 GPUs. The rest of the hyperparameters are the same as for RoBERTa~\cite{liu2019roberta}. Note that training CDLM using the large version of the Longformer model might require 2-3 times more memory and time.

\section{Finetuning on Downstream Tasks}

In this section, we elaborate further implementation details regarding the downstream tasks that we experimented, including the hyperparameter choices and the algorithms used.
\subsection{Cross-Document Coreference Resolution}
\label{subsec:ft_coref}
The setup for our cross-document coreference resolution pairwise scoring is illustrated in Figure~\ref{fig:cdcoref}. We concatenate the relevant documents using the special document separator tokens, then encode them using our \textsc{CDLM} along with the {\tt [CLS]} token at the beginning of this sequence, as suggested in Section~\ref{sec:cdmlm}. For within-document coreference candidate examples, we use just the single containing document with one set of document separators, for the single input document. Inspired by \citet{yu2020paired}, we use candidate mention marking: we wrap the mentions with special tokens $\langle$\texttt{m}$\rangle$ and $\langle$\texttt{/m}$\rangle$ in order to direct the model to specifically pay attention to the candidates representations. Additionally, we assign global-attention to {\tt [CLS]}, {\tt $\langle$\texttt{m}$\rangle$}, $\langle$\texttt{/m}$\rangle$, and the mention tokens themselves, according to the finetuning strategy proposed in Section~\ref{sec:cdmlm}. Our final pairwise-mention representation is formed like in~\citet{zeng-etal-2020-event} and~\citet{yu2020paired}: We concatenate the cross-document contextualized representation vectors for the $t^{\text{th}}$ sample:
\begin{equation*}
\begin{split}
{m_t(i,j)=\left[s_t,m_t^i,m_t^j,m_t^i\circ m_t^j \right]},
\end{split}
\end{equation*}
where $[\cdot]$ denotes the concatenation operator, $s_t$ is the cross-document contextualized representation vector of the {\tt [CLS]} token, and each of $m_t^i$ and $m_t^j$ is the sum of candidate tokens of the corresponding mentions ($i$ and $j$). Then, we train the pairwise scorer according to the suggested finetuning scheme. At test time, similar to most recent works, we apply agglomerative clustering to merge the most similar cluster pairs.

Regarding the training data collection and hyperparameter setting, we adopt the same protocol as suggested in~\citet{Cattan2020StreamliningCC}:\footnote{We used the implementation taken from \url{https://github.com/ariecattan/cross_encoder}} Our training set is composed of positive instances which consist of all the pairs of mentions that belong to the same coreference cluster, while the negative examples are randomly sampled. 

The resulting feature vector is passed through a MLP pairwise scorer that is composed of one hidden layer of the size of 1024, followed by the Tanh activation. We finetune our models for 10 epochs, with an effective batch size of 128. We used eight 32GB V100-SMX2 GPUs for finetuning our models. The finetuning process took $\sim$28 and $\sim$45 hours per epoch, for event coreference and entity coreference, respectively.

\subsection{Multi-Document Classification}
\label{subsec:app_aan}
We tune our models for 8 epochs, using a batch size of 32, and used the same hyperparameter setting from \citet[Section~5.2]{zhou-etal-2020-multilevel}.\footnote{we used the script \url{https://github.com/XuhuiZhou/CDA/blob/master/BERT-HAN/run_ex_sent.sh}} We used eight 32GB V100-SMX2 GPUs for finetuning our models. The finetuning process took $\sim$2,$\sim$5,$\sim$3, and $\sim$0.5 hours per epoch, for AAN, OC, S2ORC, and for PAN, respectively. We used the mixed-precision training mode, to reduce time and memory consumption.

\subsection{Multihop Question Answering}
\label{subsec:multihopqa}
For preparing the data for training and evaluation, we follow our finetuning scheme: for each example, we concatenate the question and all the 10 paragraphs in one long context. We particularly use the following input format with special tokens and our document separators: ``\texttt{[CLS] [q] question [/q] $\langle$\texttt{doc-s}$\rangle$$\langle$t$\rangle$ $\texttt{title}_{\texttt{1}}$ $\langle$/t$\rangle$} \texttt{$\langle$s$\rangle$} $\texttt{sent}_{\texttt{1,1}}$ \texttt{$\langle$/s$\rangle$} \texttt{$\langle$s$\rangle$} $\texttt{sent}_{\texttt{1,2}}$ \texttt{$\langle$/s$\rangle$} $\langle$\texttt{/doc-s}$\rangle$ \texttt{...} \texttt{$\langle$t$\rangle$ $\langle$\texttt{doc-s}$\rangle$ $\texttt{title}_{\texttt{2}}$ $\langle$/t$\rangle$ }  $\texttt{sent}_{\texttt{2,1}}$ \texttt{$\langle$/s$\rangle$} \texttt{$\langle$s$\rangle$} $\texttt{sent}_{\texttt{2,2}}$ \texttt{$\langle$/s$\rangle$} \texttt{$\langle$s$\rangle$} \texttt{...}'' where \texttt{[q]}, \texttt{[/q]}, $\langle$\texttt{t}$\rangle$, $\langle$\texttt{/t}$\rangle$, \texttt{$\langle$s$\rangle$}, \texttt{$\langle$/s$\rangle$}, \texttt{[p]} are special tokens representing, question start and end, paragraph title start and end, and sentence start and end, respectively. The new special tokens were added to the models vocabulary and randomly initialized before task finetuning. We use global attention to question tokens, paragraph title start tokens as well as sentence tokens. 
The model's structure is taken from ~\citet{beltagy2020longformer}.

Similar to~\citet{beltagy2020longformer}, we finetune our models for 5 epochs, using a batch size of 32, learning rate of 1e-4, 100 warmup steps. Finetuning on our models took $\sim$6 hours per epoch, using four 48GB RTX8000 GPUs for finetuning our models.

\end{document}